\documentclass[10pt,twocolumn,letterpaper]{article}

\usepackage{cvpr}              

\usepackage{graphicx} 
\usepackage{amsmath}
\usepackage{amssymb}
\usepackage{booktabs}
\usepackage{enumitem}
\usepackage{bm}
\usepackage{pifont}
\usepackage{color}
\usepackage{colortbl}
\usepackage{multirow}

\newcommand{\cmark}{\ding{51}}%
\newcommand{\xmark}{\ding{55}}%
\newcommand{\bx}{{\bm x}}
\newcommand{\venue}[1]{{$_{\text{#1}}$}}

\definecolor{Gray}{gray}{0.90}
\definecolor{LightCyan}{rgb}{0.82,0.82,1}

\definecolor{sh_gray}{rgb}{0.84,0.84,0.84}
\definecolor{sh_gray2}{rgb}{1,0.89,0.75}
\definecolor{color3}{rgb}{0.95,0.95,0.95}
\definecolor{color4}{rgb}{0.96,0.96,0.86}

\definecolor{grey}{rgb}{0.6, 0.6, 0.6}
\newcommand\dht[1]{\textcolor{grey}{#1}}

\usepackage[pagebackref,breaklinks,colorlinks]{hyperref}

\usepackage[capitalize]{cleveref}
\crefname{section}{Sec.}{Secs.}
\Crefname{section}{Section}{Sections}
\Crefname{table}{Table}{Tables}
\crefname{table}{Tab.}{Tabs.}

\def\cvprPaperID{00242} 
\def\confName{CVPR}
\def\confYear{2022}

\begin{document}

\title{Self-supervised Video Transformer}

\author{%
  Kanchana Ranasinghe$^{1,3}$ \quad 
  Muzammal Naseer$^{2,3}$ \quad 
  Salman Khan$^{3,2}$ \\
  Fahad Shahbaz Khan$^{3,4}$ \quad
  Michael Ryoo$^{1}$
  \vspace{0.5em} \\
  $^{1}$Stony Brook University \quad 
  $^{2}$Australian National University \\ 
  $^{3}$Mohamed bin Zayed University of AI \quad 
  $^{4}$Link\"{o}ping University 
  \vspace{0.1em} \\
  \small{\texttt{kranasinghe@cs.stonybrook.edu}}
}


\maketitle

\begin{abstract}

In this paper, we propose self-supervised training for video transformers using unlabeled video data. 
From a given video, we create local and global spatiotemporal views with varying spatial sizes and frame rates.
Our self-supervised objective seeks to match the features of these different views representing the same video, to be invariant to spatiotemporal variations in actions. 
To the best of our knowledge, the proposed approach is the first to alleviate the dependency on negative samples or dedicated memory banks in Self-supervised Video Transformer (SVT).
Further, owing to the flexibility of Transformer models, SVT supports slow-fast video processing within a single architecture using dynamically adjusted positional encoding and supports long-term relationship modeling along spatiotemporal dimensions. 
Our approach performs well on four action recognition benchmarks (Kinetics-400, UCF-101, HMDB-51, and SSv2) and converges faster with small batch sizes. Code is available at: \url{https://git.io/J1juJ}

\end{abstract}

\section{Introduction}
\label{sec:intro}

Self-supervised learning  enables extraction of meaningful representations from unlabeled data, alleviating the need for expensive annotations. Recent self-supervised methods perform on-par with supervised learning for certain vision tasks \cite{he2020momentum, chen2020simple, caron2018DeepCF, caron2020unsupervised}. The necessity of self-supervised learning is even greater in domains such as video analysis where annotations are more expensive \cite{piergiovanni2020evolving, recasens2021broaden, Jenni_2021_ICCV, Huang_2021_ICCV}.

At the same time, the emergence of vision transformers (ViTs) \cite{dosovitskiy2020image} and their successful adoption to different computer vision tasks including video understanding \cite{gberta_2021_ICML, liu2021video, Ryoo2021TokenLearnerWC, sharir2021image, arnab2021vivit, fan2021multiscale} within the supervised setting shows their promise in the video domain. In fact, recent works using simple ViT backbones \cite{gberta_2021_ICML} surpass convolutional neural networks (CNN) for supervised video analysis with reduced compute. Motivated by the ability of self-attention to model long-range dependencies, we propose a simple yet effective method to train video transformers \cite{gberta_2021_ICML} in a self-supervised manner. This process uses spatial and temporal context as a supervisory signal (from unlabelled videos) to learn motion, scale, and viewpoint invariant features. 

\begin{figure}[t]
    \centering
    \includegraphics[width=\linewidth]{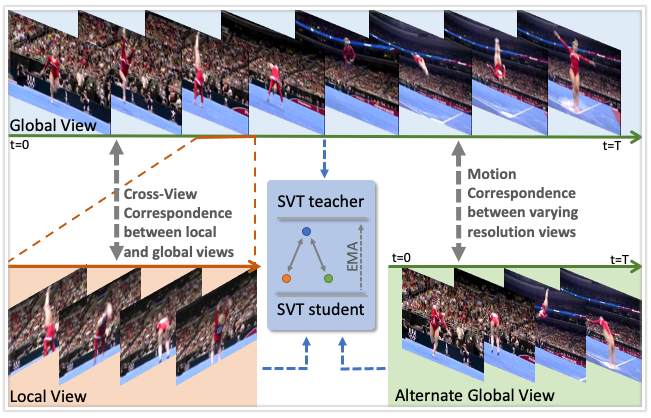}
    \caption{Our Self-supervised Video Transformer (SVT) learns cross-view and motion correspondences by jointly matching video clips sampled with varying field of view and temporal resolutions. Specifically, \emph{Global} views (top and bottom right) with different temporal resolutions as well as \emph{Local} views (bottom left) from different spatiotemporal windows are sampled. The representations of these multiple views are matched in a student-teacher framework to learn cross-view and motion correspondences (middle block). The proposed self-supervised framework can learn high-quality spatiotemporal features while converging faster.}
    \label{fig:intro}
    \vspace{-1em}
\end{figure} 

Many existing self-supervised representation learning methods on videos \cite{xie2016unsupervised, recasens2021broaden, qian2020spatiotemporal}  use contrastive learning objectives which can require larger batch sizes, longer training regimes, careful negative mining and dedicated memory banks. Further, the contrastive objectives require careful temporal sampling \cite{qian2020spatiotemporal} and multiple networks looking at similar/different clips to develop attract/repel loss formulations \cite{recasens2021broaden}. In contrast, we propose to learn self-supervised features from unlabelled videos via self-distillation \cite{tarvainen2017mean} by a twin network strategy (student-teacher models) \cite{grill2020bootstrap, caron2021emerging}.  

Our proposed approach, Self-supervised Video Transformer (SVT), trains student and teacher models with a similarity objective \cite{caron2021emerging} that matches the representations along spatial and temporal dimensions by space and time attention \cite{gberta_2021_ICML}. We achieve this by creating spatiotemporal positive views that differ in spatial sizes and are sampled at different time frames from a single video (\cref{fig:intro}). 
During training, teacher video transformer parameters are updated as an exponential moving average of the student video transformer. Both of these networks process different spatiotemporal views of the same video and our objective function is designed to predict one view from the other in the feature space. 
This allows SVT to learn robust features that are \emph{invariant} to spatiotemporal changes in videos while generating discriminative features across videos \cite{grill2020bootstrap}. SVT does not depend on negative mining or large batch sizes and remains computationally efficient as it converges within only a few epochs ($\approx$ 20 on Kinetics-400 \cite{kinetics400}). 

In addition to the above advantages, our design allows the flexibility to model varying time-resolutions and spatial scales within a unified architecture. This is a much desired feature for video processing since real-world actions can occur with varying temporal and spatial details. Remarkably, current self-supervision  based video frameworks \cite{xiao2021modist,qian2020spatiotemporal} operate on fixed spatial and temporal scales which can pose difficulties in modeling the expressivity and dynamic nature of actions. We note that convolutional backbones used in these approaches lack the adaptability to varying temporal resolutions (due to fixed number of channels) and thus require dedicated networks for each resolution \cite{feichtenhofer2019slowfast, kahatapitiya2021coarse}. To address this challenge, the proposed SVT uses dynamically adjusted positional encodings to handle varying temporal resolutions within the same architecture.  
Further, the self-attention mechanism in SVT can capture both local and global long-range dependencies across both space and time, offering much larger receptive fields as compared to traditional convolutional kernels \cite{naseer2021intriguing}.

\noindent The main contributions in this work are as follows:\vspace{-0.5em} 
\begin{itemize}[leftmargin=*,noitemsep, topsep=1.0ex,itemsep=-0.5ex,partopsep=0ex,parsep=1ex]
    \item 	We introduce a novel mechanism for self-supervised training of video transformers	by exploiting spatiotemporal correspondences between 
	varying fields of view (global and local) across space and time (\cref{subsec:training}).   
	\item Self-supervision in SVT is performed via a joint motion and crossview correspondence learning objective. 
	Specifically, global and local spatiotemporal views with varying frame rates and spatial characteristics (\cref{subsec:mo_pred}  \cref{subsec:cross_view_corr}) are matched by our motion and crossview correspondences in the latent space.
	\item 
	A unique property of our architecture is that it allows slow-fast training and inference using a single video transformer. To this end, we propose to use dynamic positional encoding within SVT to handle variable frame rate inputs generated from our sampling strategy (\cref{subsec:dynamic_positional_embedding}). 
\end{itemize}

Our extensive experiments and results on various video datasets including Kinetics-400 \cite{kinetics400}, UCF-101 \cite{soomro2012ucf}, HMDB-51 \cite{kuehne2011hmdb}, and SSv2 \cite{goyal2017something} show state-of-the-art transfer of our self-supervised features using only RGB data. Further, our method shows a rapid convergence rate.
\section{Related Work}
\label{sec:related}

\begin{figure*}[t]
	\centering
	\includegraphics[width=\linewidth]{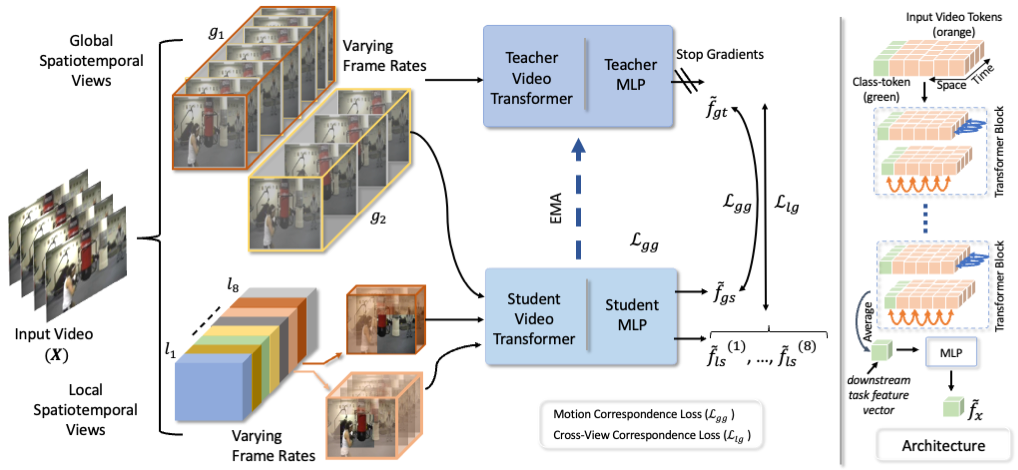}
	\vspace{-2em}
	\caption{Our spatiotemporal sampling generates global and local views from a given input video. Global views contain different frame rates and spatial characteristics controlled by sampling strategy and combinations of augmentations. Local views have varying frame rates as well as limited fields of view due to random cropping. One global view is randomly selected and passed through the teacher model to generate a target while other global and local views are passed through the student model. Network weights are then updated by matching the online student local (\emph{cross-view correspondences}) and global (\emph{motion correspondences}) views to the target teacher global view. We use a standard ViT backbone with separate space-time attention \cite{gberta_2021_ICML} followed by an MLP for predicting target features from online features.
	}
	\label{fig:architecture}
	\vspace{-1em}
\end{figure*}


\textbf{Transformers in Vision.}
Since the initial success of transformers in natural language processing (NLP) tasks \cite{vaswani2017attention, devlin2018bert}, they have emerged as a competitive architecture for various other domains \cite{khan2021transformers}. Among vision tasks, the initial works focused on a combination of convolutional and self-attention based architectures \cite{carion2020end, wang2018non, wang2020end, zhang2020dynamic}. A convolution free variant, vision transformer (ViT) \cite{dosovitskiy2020image}, achieved competitive performance on image classification tasks. While earlier works proposing ViT \cite{dosovitskiy2020image} depended on large-scale datasets, more recent efforts achieve similar results with medium-scale datasets using various augmentation strategies \cite{pmlr-v139-touvron21a, Steiner2021HowTT}. Later architectures also explore improving computational efficiency of ViTs focusing on transformer blocks \cite{liu2021swin, Ryoo2021TokenLearnerWC}. ViTs have also been adopted for video classification tasks \cite{gberta_2021_ICML, sharir2021image, arnab2021vivit, fan2021multiscale, Ryoo2021TokenLearnerWC}. Our work builds on the TimeSformer backbone \cite{gberta_2021_ICML}, a direct adaptation of standard ViTs using separate attention across dimensions.

\textbf{Self-supervised Learning in Images.}
Early image-based self-supervised learning work focused on pretext tasks that require useful representations to solve \cite{komodakis2018unsupervised, pathak2016context, vincent2008extracting, noroozi2016unsupervised, doersch2015unsupervised, doersch2017multi, zhang2016colorful}. However, recently contrastive methods have dominated self-supervised learning \cite{caron2021emerging, chen2020simple, chen2020improved, he2020momentum, chen2021empirical, AMDIM, dosovitskiy2014discriminative, henaff2019data, le2020contrastive, PIRL, tian2019contrastive, tian2020makes} . These approaches generally consider two views of a single sample (transformed through augmentations) and pull them (positives) together while pushing away from all other (negative) samples in representation space \cite{AMDIM, oord2018representation}. Key drawbacks of these methods are the necessity for careful mining of positive / negative samples \cite{tian2020makes} and reliance on large numbers of negative samples (leading to large batch sizes \cite{chen2020simple} or memory banks \cite{he2020momentum}).  While clustering methods improve on this using cluster targets \cite{xie2016unsupervised, tian2017deepcluster, bautista2016cliquecnn, caron2018DeepCF, caron2020unsupervised, alwassel2019self, asano2020selflabelling, huang2019unsupervised}, recent regression based methods that predict alternate representations \cite{grill2020bootstrap, caron2021emerging} eliminate the need for sample mining and negative samples. In particular, Caron \etal \cite{caron2021emerging} explore predicting spatially local-global correspondences with ViT backbones within the image domain, which we extend in our work to the video domain with suitable improvements. 

\textbf{Self-supervised Learning in Videos.}
While self-supervised learning in videos were initially dominated by approaches based on pretext tasks unique to the video domain \cite{mathieu2015deep, PatrauceanHC16, walker2016uncertain, pmlr-v37-srivastava15, Vondrick16a, vondrick2018tracking, Agrawal_2015_ICCV, Goroshin_2015_ICCV, DBLP:journals/corr/IsolaZKA15, Misra-2016-5596, 7410677}, recent work focuses more on contrastive losses similar to the image domain \cite{Feichtenhofer_large, han2019video, han2020self, qian2020spatiotemporal, hjelm2020representation, recasens2021broaden}. 
A combination of previous pretext tasks over multiple modalities with cross-modality distillation is presented in \cite{piergiovanni2020evolving};  SVT differs in how our self-distillation operates within a single modality and network.
The idea of varying resolution along temporal dimension is explored in \cite{Huang_2021_ICCV, chen2021rspnet}. They use contrastive losses between different videos at same resolution for speed consistency or the same video at different resolutions for appearance consistency. Unlike these works, we jointly vary spatial and temporal resolutions and use a predictive objective as self-supervision.  The idea of views with limited locality is also explored in \cite{Behrmann2021LongSV, recasens2021broaden, Dave2021TCLRTC}. While \cite{Behrmann2021LongSV} uses views of varying locality for disentangling the representation space into temporally local and global features using contrastive objectives, our approach uses view locality to learn correspondences along and across dimensions with our predictive objective. A similar predictive objective with temporal locality constrained views is used in \cite{recasens2021broaden} and contrastive losses with spatial local-global crops is used in \cite{Dave2021TCLRTC}; however our approach focuses on spatio-temporal constraints extending correspondences across dimensions, uses a single shared network for processing alternate views, and additionally combines varying resolutions to generate our alternate views exploiting unique ViT architectural features.

\section{Self-supervised Video Transformer}
\label{sec:method}

In this section, we discuss our Self-supervised Video Transformer (SVT) approach. 
Unlike contrastive methods, we process two clips from the same video with varying spatial-temporal characteristics (\cref{subsec:training}) avoiding the need for negative mining or memory banks. 
Our loss formulation matches the representations from both dissimilar clips to enforce invariance to motion and spatial changes for the same action sequence. 
A naive objective enforcing invariance would collapse all representations to be the same, however we use a teacher-student network pair where the former acts as a more stable target for the later, enabling convergence of the online student network to learn discriminative representations \cite{grill2020bootstrap}.
This approach simultaneously incorporates rich spatiotemporal context in the representations while keeping them discriminative. In the following, we first introduce the overall architecture of SVT in \cref{subsec:arch} followed by the self-supervised learning process in \cref{subsec:training}, our objective functions in \cref{subsec:svt_loss} and inference in \cref{subsec:sf_inf}.

\subsection{SVT: Architecture}\label{subsec:arch}
We apply separate attention along temporal and spatial dimensions of input video clips using a video transformer \cite{gberta_2021_ICML}. 
Consider a video $\bm{X}=\{\bx_t\}_{t=1}^N$, where $N$ represents the number of frames. 
We define a \textit{clip} (also termed \textit{view} interchangeably) as a subset of these $N$ frames selected through a sampling strategy. 
We define $H$, $W$, $T$ to be the height, width, and number of frames respectively for the sampled clip. Our sampling methodology (\cref{fig:architecture}) generates two types of clips, \textit{global} ($\bm{g}$) and \textit{local} ($\bm{l}$) spatiotemporal views. Both $\bm{g}$ and $\bm{l}$ are subsets of the video frame set $\bm{X}$, with views $\bm{g}=\{\bx'_t\}_{t=1}^{K_{g}}$, $ \bm{l}=\{\bx''_t\}_{t=1}^{K_{l}}$, and $|K_{l}| \le |K_{g}|$. 
\noindent \textbf{Global views} are generated by uniformly sampling a variable number of frames from a randomly selected 90\% portion of a video's time axis. We generate two such global spatiotemporal views ($\bm{g}_1,\bm{g}_2$) at low ($T=8$) and high ($T=16$) frame rates and spatial resolution $H=W=224$. \\
\noindent \textbf{Local views} are generated by uniformly sampling frames from a randomly selected video region covering $1/8^{th}$ of the time axis and $\approx$ 40\% area along spatial axes. We generate eight such local spatiotemporal views with $T\in\{2,4,8,16\}$ and spatial resolution fixed at $H=W=96$. 
Specifically, we randomly sample two global ($\bm{g}_1,\bm{g}_2$) and eight local ($\bm{l}_1,\ldots,\bm{l}_8$) spatiotemporal views. Note that both spatial and temporal dimensions within our sampled views differ from those of the original video.
We introduce the channel dimension, $C$, which is fixed at 3 for RGB inputs considered in our case. Our SVT, comprising of 12 encoder blocks, processes each sampled clip of shape $(C\times T\times W\times H)$, where $W\le 224$, $H\le 224$ and $T\le 16$ (different for each clip).
Our network architecture (\cref{fig:architecture}) is designed to process such varied resolution clips during both training and inference stages within a single architecture (Sec.~\ref{subsec:dynamic_positional_embedding}). 

During training, we divide each frame within a view into patches \cite{dosovitskiy2020image}. 
Thus, for a given view of maximum size $H=W=224$ and $T=16$, each SVT encoder block processes a maximum of 196 spatial and 16 temporal tokens, and the embedding dimension of each token is $\mathbb{R}^{768}$ \cite{dosovitskiy2020image}. 
Since the maximum number of spatial and temporal tokens vary due to variable dimensions in our views, we deploy {dynamic positional encoding (\cref{subsec:dynamic_positional_embedding})} to account for any missing tokens for views of size $W< 224$, $H< 224$ and $T< 16$. Note the minimum spatial and temporal sizes in our proposed views are $H=W=96$ and $T=2$, respectively.
In addition to these input spatial and temporal tokens, we use a single classification token as the feature vector within the architecture \cite{devlin2018bert, dosovitskiy2020image}. This classification token represents the common features learned by the SVT along spatial and temporal dimensions of a given video. Finally, we use a multi-layer perceptron (MLP) as a projection head over the classification token from the final encoder block \cite{caron2021emerging, grill2020bootstrap}. We define the output of our projection head as $\bm{f}$. 

As illustrated in \cref{fig:architecture}, our overall approach uses a teacher-student setup inspired from \cite{caron2021emerging, grill2020bootstrap} for self-distillation \cite{tarvainen2017mean}. 
Our teacher model is an exact architectural replica of the student model.

\subsection{SVT: Self-supervised Training}
\label{subsec:training}

We train SVT in a self-supervised manner by predicting the different views (video clips) with varying spatiotemporal characteristics from each other in the feature space of student and teacher models. To this end, we adopt a simple routing strategy that randomly selects and passes different views through the teacher and student models. The teacher SVT processes a given global spatiotemporal view to produce a feature vector, $\bm{f}_{g_t}$, which is used as the target label, while the student SVT processes local and global spatiotemporal views to produce feature vectors, $\bm{f}_{g_s}$, and $\bm{f}^{(1)}_{l_s}, ..., \bm{f}^{(8)}_{l_s}$, which are matched to the target feature $\bm{f}_{g_t}$ through our proposed loss (Eq.~\ref{eq:svt_loss}). During each training step, we update the student model weights via backpropagation while teacher weights are updated as an exponential moving average (EMA) of the student weights \cite{caron2021emerging}. 

Our motivation for predicting such varying views of a video is that it leads to modeling the contextual information defining the underlying distribution of videos by learning motion correspondences (global to global spatiotemporal view matching) and cross-view correspondences (local to global spatiotemporal view matching) (\cref{fig:attn_vis}).  This makes the model invariant to motion, scale and viewpoint variations. Thus, our self-supervised video representation learning approach depends on closing the gap between feature representations of different spatiotemporal views from the same video using a self-distillation mechanism. Next, we explain how motion correspondences and cross-view correspondences are learned, followed by our loss formulation. 

\vspace{-0.7em}
\subsubsection{Motion Correspondences}
\label{subsec:mo_pred}
\vspace{-0.5em}
A defining characteristic of a video is the frame rate. Varying the frame rate can change motion context of a video (\eg, walking slow vs walking fast) while controlling nuanced actions (\eg, subtle body-movements of walking action). In general, clips are sampled from videos at a fixed frame rate \cite{qian2020spatiotemporal, xiao2021modist}. However, given two clips of varying frame rate (different number of total frames for each clip), predicting one from the other in feature space explicitly involves modeling the motion correspondences (MC) of objects across frames. Further, predicting subtle movements captured at high frame rates will force a model to learn motion related contextual information from a low frame rate input. We model this desired property into our training by matching global to global spatiotemporal views. Refer to \Cref{app:view_routes} for further details. 

\vspace{-0.7em}
\subsubsection{Cross-View Correspondences}
\label{subsec:cross_view_corr}
\vspace{-0.5em}
In addition to learning motion correspondences, our training strategy aims to model relationships across spatiotemporal variations as well by learning cross-view correspondences (CVC). The cross-view correspondences are learned by matching the local spatiotemporal views processed by our student SVT ($\bm{f}^{(i)}_{l_s}: i\in [1,8]$) with a global spatiotemporal view representation processed by our teacher SVT model ($\bm{f}_{g_t}$). Our local views cover a limited portion of videos along both spatial and temporal dimensions. 

Our intuition is that predicting a global spatiotemporal view of a video from a local spatiotemporal view in the latent space forces the model to learn high-level contextual information by modeling, \textbf{a)} spatial context in the form of possible neighbourhoods of a given spatial crop, and \textbf{b)} temporal context in the form of possible previous or future frames from a given temporal crop. Note that in the cross-view correspondences, we predict a global view frame using all frames of a local view by our similarity objective (Eq.~\ref{eq:local_to_global_loss}). 

%
%

\vspace{-0.7em}
\subsubsection{Dynamic Positional Embedding}
\vspace{-0.5em}
\label{subsec:dynamic_positional_embedding}
Vision transformers \cite{dosovitskiy2020image} require inputs to be converted to sequences of tokens, which allows efficient parallel processing. Positional encoding is used to model ordering of these sequences \cite{naseer2021intriguing}. Interestingly, positional encoding also allows ViT to process variable input resolution by interpolating the positional embedding for the missing tokens. As mentioned earlier, our motion and cross-view correspondences involve varying spatial and temporal resolutions which results in variable spatial and temporal input tokens during training (\cref{subsec:arch}). We use this property of positional encoding to our advantage by accommodating varying spatial and temporal tokens in our proposed training mechanism. 
In implementing this, during training we use a separate positional encoding vector for spatial and temporal dimensions and fix these vectors to the highest resolution across each dimension. Similar to \cite{dosovitskiy2020image}, our positional encoding is a learned vector. We vary the positional encoding vectors through interpolation during training to account for the missing spatial or temporal tokens at lower frame rate or spatial size. This allows our single SVT model to process inputs of varying resolution while also giving the positional embedding a dynamic nature which is more suited for different sized inputs in the downstream tasks. During slow-fast inference (\Cref{subsec:sf_inf}) on downstream tasks, the positional encoding is interpolated to the maximum frame count and spatial resolution used across all views. 

We note that our learned positional encoding is implicitly tied to frame number to cue the relative ordering of the sampled frames. Given the varying frame rates of views, it does not encode the exact time stamp (frame rate information). We hypothesize that despite not differentiating frame rates, cuing frame order is sufficient for SVT training. 

\vspace{-0.7em}
\subsubsection{Augmentations}
\vspace{-0.5em}
In addition to our sampling strategy (temporal dimension augmentations), we also apply standard image augmentations to the spatial dimension, \ie, augmentations are applied to the individual frames sampled for each view. We follow temporally consistent spatial augmentations \cite{qian2020spatiotemporal} where the same randomly selected augmentation is applied equally to all frames belonging to a single view. The standard augmentations used include random color jitter, gray scaling, Gaussian blur, and solarization. We also apply random horizontal flips to datasets not containing flip equivariant classes (\eg, walking left to right).

\subsection{SVT Loss}
\label{subsec:svt_loss}
We enforce motion and cross-view correspondences by matching our proposed spatiotemporal views within the feature space. Specifically, we match global to global views to learn motion and local to global views to learn cross-view correspondences by minimizing the following objective:
\begin{align}\label{eq:svt_loss}
    \mathcal{L} =  \mathcal{L}_{lg} +  \mathcal{L}_{gg}.
\end{align}
The global and local spatiotemporal views are passed though the student and teacher models to get the corresponding feature outputs $\bm{f}_g$ and $\bm{f}_l$. These feature vectors are normalized to obtain $\bm{\Tilde{f}}$ as follows: 
\begin{equation}
     \bm{\Tilde{f}}{[i]} = \frac{\text{exp}(\bm{f}{[i]}) / \tau}{\sum_{i=1}^{n} \text{exp}(\bm{f}{[i]})/ \tau },\notag
\end{equation}
where $\tau$ is a temperature parameter used to control sharpness of the exponential function \cite{caron2021emerging} and $\bm{\Tilde{f}}{[i]}$ is each element of $\bm{\Tilde{f}} \in \mathbb{R}^{n}$.  

\vspace{0.1em}
\noindent\textbf{Motion Correspondence Loss:}
We forward pass a global view through the teacher SVT serving as the target feature which is compared with an alternate global view processed by the student SVT to obtain a loss term (Eq.~\ref{eq:global_to_global_loss}). This loss measures the difference in motion correspondences between these two global views.
\begin{align}\label{eq:global_to_global_loss}
    \mathcal{L}_{gg} &=  -\bm{\Tilde{f}}_{g_{t}} \cdot \log(\bm{\Tilde{f}}_{g_{s}}),
\end{align}
where, $\bm{\Tilde{f}}_{g_{s}}$ and $\bm{\Tilde{f}}_{g_{t}}$ are the feature outputs of different global spatiotemporal views from the student and teacher network respectively and $[\cdot]$ is dot product operator.

\vspace{0.1em}
\noindent\textbf{Cross-view Correspondence Loss:} 
All local spatiotemporal views are passed through the student SVT model and mapped to a global spatiotemporal view from the teacher SVT model to reduce the difference in feature representation, learning cross-view correspondences (Eq.~\ref{eq:local_to_global_loss}).
\begin{align}\label{eq:local_to_global_loss}
    \mathcal{L}_{lg} &= \sum_{i=1}^{k} -\bm{\Tilde{f}}_{g_{t}} \cdot \log(\bm{\Tilde{f}}^{(i)}_{l_{s}}),
\end{align}
where the sum is performed over $k$ different local spatiotemporal views ($k=8$ used consistently across all experiments) and $\bm{\Tilde{f}}^{(i)}_{l_{s}}$ are the feature outputs for $i^{th}$ local view. 

\vspace{0.1em}
\noindent\textbf{Convergence:} 
Given our two separate student ($\theta$) and teacher ($\xi$) networks, let us view our overall loss, $L$ as a function of their learnable parameters, $L_{\theta, \xi}$. There exists a concern of collapse to a trivial solution (teacher and student outputs always equal a constant) during training. However, we note that SVT parameters do not converge to such a minimum over $L_{\theta, \xi}$ because: \textbf{a)} The SVT teacher parameter updates are not in the direction of $\nabla_{\xi} L_{\theta, \xi}$ since 
$ \xi_{t+1} \leftarrow \tau \xi_t + (1-\tau) \theta_t $ for $\tau \in [0,1]$ (EMA update).
\textbf{b)} SVT's gradient descent on $L_{\theta, \xi}$ does not act jointly over $(\theta, \xi)$. 
This is similar to BYOL \cite{grill2020bootstrap} where such a loss acts on the outputs of student and teacher networks. Additionally, as suggested in \cite{caron2021emerging}, we also use centering and sharpening of teacher outputs to further facilitate convergence. 

\vspace{0.5em}
\subsection{SVT: Slow-Fast Inference}
\label{subsec:sf_inf}
\begin{figure}[!t]
    \vspace{0.5em}
    \centering
    \includegraphics[width=0.99\linewidth]{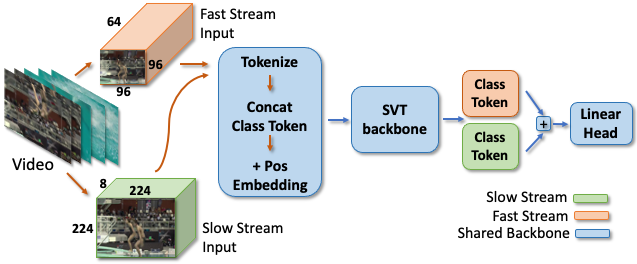}
    \vspace{-0.5em}
    \caption{Slow-Fast Inference: we uniformly sample two clips of the same video at resolutions $(8, 224, 244)$ and $(64, 96, 96)$, pass through a shared network, and generate two different feature vectors (class tokens). These vectors are combined in a deterministic manner (with no learnable parameters), e.g. summation, to obtain a joint vector that is fed to the downstream task classifier.
    }
    \label{fig:slowfast}
    \vspace{-1em}
\end{figure}

Slow-Fast inference refers to using two different video clips with \textit{high spatial but low temporal} and \textit{low spatial but high temporal} resolutions. This allows capturing finer-information across each dimension with minimal increase in computation. Recent methods \cite{feichtenhofer2019slowfast, kahatapitiya2021coarse} deploy such inference but use multiple network architectures for processing videos at different resolutions. However, our dynamic positional encodings allow Slow-Fast inference within our single SVT model (Sec.~\ref{subsec:dynamic_positional_embedding}) as illustrated in \Cref{fig:slowfast}. We use this on downstream tasks for improved performance.

\section{Experiments}
\label{sec:experiments}

\subsection{Experimental Setup and Protocols}
\label{subsec:exp_setup}

\noindent\textbf{Datasets:} 
We use the Kinetics-400 data \cite{kinetics400} (train set) for the self-supervised training phase of SVT. We use its validation set for evaluation. Additionally, we evaluate on three downstream datasets, UCF-101 \cite{soomro2012ucf}, HMBD-51 \cite{kuehne2011hmdb}, and Something-Something v2 (SSv2) \cite{goyal2017something}. 

\begin{table*}[t]
\centering
\small
\caption{\textbf{UCF-101 \cite{soomro2012ucf} \& HMBD-51 \cite{kuehne2011hmdb}:} Top-1 (\%) accuracy for both linear evaluation and fine-tuning. All models are pre-trained on Kinetics-400 \cite{kinetics400} except ELo \cite{piergiovanni2020evolving} which uses YouTube8M dataset \cite{youtube8M}. Gray shaded methods use additional optical flow inputs. S-Res and T-Res represent spatial and temporal input resolution, respectively. Our approach shows state-of-the-art or on par performance.
}
\begin{tabular}{l|c|c|c|c|c|cc|cc} %
\rowcolor{Gray}
\toprule
                                                      &           &   &        &  &        & \multicolumn{2}{c|}{UCF-101 \cite{soomro2012ucf}}          & \multicolumn{2}{c}{HMDB-51 \cite{kuehne2011hmdb}}          \\ \cline{7-10}
\rowcolor{Gray}                                                             
\multirow{-2}{*}{Method} & \multirow{-2}{*}{Backbone}  & \multirow{-2}{*}{TFLOPs} & \multirow{-2}{*}{S-Res}  & \multirow{-2}{*}{T-Res}  & \multirow{-2}{*}{Epochs}    & \multicolumn{1}{c|}{Linear}& Fine-tune   & \multicolumn{1}{c|}{Linear}& Fine-tune   \\ \midrule
MemDPC \cite{han2019video} \venue{(ECCV ‘20)}         & R2D3D-34  & -     & 224    & 64   & -   & \multicolumn{1}{c|}{54.1}  & 86.1 & \multicolumn{1}{c|}{30.5}  & 54.5 \\ 
CoCLR \cite{han2020self} \venue{(Neurips ‘20)}        & S3D       & 0.07  & 128    & 32   & 100  & \multicolumn{1}{c|}{77.8}  & 87.9 & \multicolumn{1}{c|}{52.4}  & 54.6 \\ 
ELo \cite{piergiovanni2020evolving} \venue{(CVPR ’20)}& R(2+1)D   & 17.5  & 224    &  -   & 100  & \multicolumn{1}{c|}{-}     & 84.2 & \multicolumn{1}{c|}{-}     & 53.7 \\ 
RSPNet \cite{chen2021rspnet} \venue{(AAAI '21)}       & S3D-G     & 0.07  & 112    & 16   & 200  & \multicolumn{1}{c|}{-}     & 89.9 & \multicolumn{1}{c|}{-}     & 59.6 \\ 
VideoMoCo \cite{pan2021videomoco}\venue{(CVPR ‘21)}   & R(2+1)D   & 17.5  & 112    & 32   & 200  & \multicolumn{1}{c|}{78.7}  & -    & \multicolumn{1}{c|}{49.2}  & -    \\ 
BE \cite{wang2021removing} \venue{(CVPR ‘21)}         & I3D       & 2.22  & 224    & 16   & 50   & \multicolumn{1}{c|}{-}     & 87.1 & \multicolumn{1}{c|}{-}     & 56.2 \\
CMD \cite{Huang2021SSVR}\venue{(CVPR ‘21)}            & R(2+1)D-26& -     & 112    & 16   & 120  & \multicolumn{1}{c|}{-}     & 85.7 & \multicolumn{1}{c|}{-}     & 54.0 \\ 
CVRL \cite{qian2020spatiotemporal}\venue{(CVPR ‘21)}  & R3D-50    & 3.19  & 224    & 32   & 800  & \multicolumn{1}{c|}{89.2}  & 92.2 & \multicolumn{1}{c|}{57.3}  & 66.7 \\ 
TCLR \cite{Dave2021TCLRTC} \venue{(Arxiv ’21)}        &R(2+1)D-18 & -     & 112    & 16   & 100  & \multicolumn{1}{c|}{-}     & 84.3 & \multicolumn{1}{c|}{-}     & 54.2 \\
\dht{MoDist \cite{xiao2021modist} \venue{(Arxiv ’21)}} &\dht{R3D-50} &\dht{3.19}&\dht{224}&\dht{8}&\dht{100}&\multicolumn{1}{c|}{\dht{91.5}} & \dht{94.0} & \multicolumn{1}{c|}{\dht{63.0}}  & \dht{67.4}\\ 
\dht{BraVe \cite{recasens2021broaden}\venue{(ICCV '21)}}&\dht{R3D-50}&\dht{3.19}&\dht{224}&\dht{16}&\dht{-}&\multicolumn{1}{c|}{\dht{92.5}}  & \dht{95.1} & \multicolumn{1}{c|}{\dht{68.3}}  & \dht{74.6} \\  
Vi$^2$CLR \cite{Diba_2021_ICCV} \venue{(ICCV '21)}    & S3D       & 0.07  & 128    & 32   & 300  & \multicolumn{1}{c|}{75.4}  & 89.1 & \multicolumn{1}{c|}{47.3}  & 55.7 \\ 
ASCNet \cite{Huang_2021_ICCV} \venue{(ICCV '21)}      & S3D-G     & 0.07  & 224    & 64   & 200  & \multicolumn{1}{c|}{-}     & 90.8 & \multicolumn{1}{c|}{-}     & 60.5 \\ 
TEC \cite{Jenni_2021_ICCV} \venue{(ICCV '21)}         & S3D-G     & 0.07  & 128    & 32   & 200  & \multicolumn{1}{c|}{-}     & 88.2 & \multicolumn{1}{c|}{-}     & 63.5 \\
LSFD \cite{Behrmann2021LongSV} \venue{(ICCV '21)}     & C3D       & -     & 224    & 16 & -    & \multicolumn{1}{c|}{-}     & 79.8 & \multicolumn{1}{c|}{-}     & 52.1 \\ 
MCN \cite{Lin_2021_ICCV} \venue{(ICCV '21)}           & R3D       & 3.19  & 128    & 32   & 50   & \multicolumn{1}{c|}{73.1}  & 89.7 & \multicolumn{1}{c|}{42.9}  & 59.3 \\ 
CORP \cite{Hu_2021_ICCV} \venue{(ICCV '21)}           & R3D-50    & 3.19  & 224    & 16   & 800   & \multicolumn{1}{c|}{90.2}  & 93.5 & \multicolumn{1}{c|}{\textbf{58.7}}  & \textbf{68.0} \\ \midrule
SVT (Ours)                                            & ViT-B \cite{gberta_2021_ICML}       
                                                                  & 0.59  & 224    & 16   & 20   & \multicolumn{1}{c|}{\textbf{90.8}}  & \textbf{93.7} & \multicolumn{1}{c|}{57.8}  & 67.2 \\ \bottomrule
      
\end{tabular}
\label{tbl:sota_comp}
\end{table*}

\begin{table}[t]
\centering\small
\caption{\textbf{Kinetics-400 \cite{kinetics400}:} Top-1 (\%) accuracy is reported for both linear evaluation and fine-tuning on the Kinetics-400 validation set. All models are pre-trained on the training set of Kinetics-400 dataset. Our approach shows state-of-the-art performance. }
	\setlength{\tabcolsep}{4pt}
	\scalebox{1.0}[1.0]{
\begin{tabular}{l|c|c|c}
\toprule
\rowcolor{Gray} 
Method                                              & Backbone & Fine-tune   & Linear \\ \midrule
CVRL \cite{qian2020spatiotemporal} \venue{(CVPR’21)}& R3D-101  & 70.4 & 67.6  \\
Vi$^2$CLR \cite{Diba_2021_ICCV} \venue{(ICCV '21)}  & S3D     & 71.2 & 63.4  \\ 
CORP \cite{Hu_2021_ICCV}        \venue{(ICCV '21)}  & R3D-50   & -    & 66.6  \\\midrule
SVT (Ours)                                          & ViT-B \cite{gberta_2021_ICML}    & \textbf{78.1} & \textbf{68.1} \\ \bottomrule
\end{tabular}}       
\label{tbl:sota_k400}
\end{table}

\begin{table}[t]
\centering\small
\caption{\textbf{SSv2 \cite{goyal2017something}:} Top-1 (\%) for both linear evaluation and fine-tuning on the SSv2 validation set. All models are pre-trained on Kinetics-400. Our approach produces best results.}
	\setlength{\tabcolsep}{4pt}
	\scalebox{1.0}[1.0]{
\begin{tabular}{l|c|c|c}
\toprule
\rowcolor{Gray} 
Method                                           & Backbone  & Fine-tune   & Linear \\ \midrule
MoDist \cite{xiao2021modist} \venue{(Arxiv’21)}  & R3D-50        & 54.9 & 16.6  \\ 
CORP \cite{Hu_2021_ICCV}      \venue{(ICCV '21)} & R3D-50        & 48.8 & -     \\\midrule
SVT (Ours)                                       & ViT-B\cite{gberta_2021_ICML} & \textbf{59.2} & \textbf{18.3} \\ \bottomrule
\end{tabular}}       
\label{tbl:sota_ssv2}
\end{table}


\begin{table*}[t]
\begin{minipage}{.55\textwidth}
	\centering\small
    \caption{\textbf{View Correspondences.} Predicting local to global and global to global views remains optimal over any other combination.}
	\setlength{\tabcolsep}{8pt}
	\scalebox{1.0}[1.0]{
	\begin{tabular}{c|c|c|c|c|c}
		\toprule 
		\rowcolor{Gray}
		$\bm{l}\to \bm{g}$  & $\bm{g}\to\bm{g}$  & $\bm{l}\to\bm{l}$  & $\bm{g}\to\bm{l}$ & UCF-101     & HMDB-51  \\  \midrule
		\cmark   & \xmark   & \xmark   & \xmark  & 84.11   & 50.72 \\ 
		\xmark   & \cmark   & \xmark   & \xmark  & 81.95   & 49.04 \\ 
		\cmark   & \cmark   & \xmark   & \xmark  & \textbf{84.64}   & \textbf{52.17} \\
		\cmark   & \cmark   & \cmark   & \xmark  & 83.11   & 51.23 \\
		\cmark   & \cmark   & \xmark   & \cmark  & 84.71   & 51.88 \\
		\cmark   & \cmark   & \cmark   & \cmark  & 83.69   & 51.71 \\ \bottomrule
	\end{tabular}
	}
    \label{tbl:ablation_correspondences}
\end{minipage}
\hfill
\begin{minipage}{.42\textwidth}
	\centering\small
\caption{\textbf{Spatial vs Temporal variations.}  Cross-view correspondences with varying field of view along both spatial and temporal dimensions lead to optimal results. Temporal variations between views has more effect than applying only spatial variation.}
	\setlength{\tabcolsep}{7pt}
	\scalebox{1.0}[1.0]{
	\begin{tabular}{c|c|c|c}
		\toprule
		\rowcolor{Gray} 
		Spatial  & Temporal & UCF-101     & HMDB-51 \\ \midrule
		\cmark   & \xmark   & 73.81   & 42.91   \\ 
		\xmark   & \cmark   & 82.90   & 42.59   \\ 
		\cmark   & \cmark   & \textbf{84.64}   & \textbf{52.17}   \\ \bottomrule
	\end{tabular}}
    \label{tbl:ablation_st}
\end{minipage}
\end{table*}

\begin{table*}[t]
\begin{minipage}{.33\textwidth}
	\centering\small
	 \caption{\textbf{Temporal Sampling Strategy 
    }. We compare our proposed temporal sampling strategy, motion correspondences (MC) (\cref{subsec:mo_pred}), against the alternate approach of temporal interval sampler (TIS) \cite{qian2020spatiotemporal} used with CNNs under contrastive settings. 
    }
	\setlength{\tabcolsep}{5pt}
	\scalebox{1.0}[1.0]{
	\begin{tabular}{c|c|c}
		\toprule
		\rowcolor{Gray}
		      & UCF-101     & HMDB-51  \\  \midrule
		Ours + TIS \cite{qian2020spatiotemporal} & 82.24   & 50.10 \\ 
		Ours + MC & \textbf{84.64}   & \textbf{52.17} \\ \bottomrule
	\end{tabular}}
    \label{tbl:ablation_temporal_aug}
\end{minipage}
\hfill
\begin{minipage}{.30\textwidth}
	\centering\small
    \caption{\textbf{Augmentations}: Using temporally consistent augmentations (TCA) \cite{qian2020spatiotemporal} applied randomly over the spatial dimensions for different views result in consistent improvements on UCF-101 and HMDB-51 datasets. 
}
	\setlength{\tabcolsep}{5pt}
	\scalebox{0.95}[0.95]{
	\begin{tabular}{c|c|c}
		\toprule
		\rowcolor{Gray}
		          & UCF-101     & HMDB-51  \\  \midrule
		w/o  TCA \cite{qian2020spatiotemporal} & 84.20   & 52.10 \\
	    w TCA \cite{qian2020spatiotemporal}    & \textbf{84.64}   & \textbf{52.17} \\ \bottomrule
	\end{tabular}}
    \label{tbl:ablation_aug}
\end{minipage}
\hfill
\begin{minipage}{.31\textwidth}
	\centering\small
    \caption{\textbf{Slow-Fast Inference}: Feeding multiple views of varying spatiotemporal resolutions to a single shared network (multi-view) results in clear performance gains over feeding single-views across both UCF-101 and HMDB-51 datasets.}
    \label{tbl:ablation_inf}
	\setlength{\tabcolsep}{5pt}
	\scalebox{1.0}[1.0]{
	\begin{tabular}{c|c|c}
		\toprule
		\rowcolor{Gray}
		Slow-Fast       & UCF-101     & HMDB-51  \\  \midrule
		\xmark      & 84.64   & 52.17 \\
		\cmark      & \textbf{84.80}   & \textbf{53.22} \\ \bottomrule
	\end{tabular}}
\end{minipage}
\end{table*}

\vspace{0.1em}
\noindent\textbf{Self-supervised Training:} We train our models for 20 epochs on the train set of Kinetics-400 dataset \cite{kinetics400} without any labels using a batch size of 32 across 4 NVIDIA-A100 GPUs. This batch size refers to the number of unique videos present within a given batch. 
We randomly initialize weights relevant to temporal attention while spatial attention weights are initialized using a ViT model trained in a self-supervised manner over the ImageNet-1k dataset \cite{imagenet}. We follow this initialization setup to obtain faster space-time ViT convergence similar to the supervised setting \cite{gberta_2021_ICML}. We use an Adam optimizer \cite{kingma15adam} with a learning rate of $5e-4$ scaled using a cosine schedule with linear warmup for 5 epochs \cite{Steiner2021HowTT, chen2021mocov3}. We also use weight decay scaled from 0.04 to 0.1 during training. Our code builds over the training frameworks from \cite{gberta_2021_ICML, fan2020pyslowfast, rw2019timm, caron2021emerging}. 

\vspace{0.1em}
\noindent\textbf{Downstream Tasks:} We perform two types of evaluations on our downstream task of action recognition for each dataset, \textbf{a) Linear:} We train a linear classifier over our pretrained SVT backbone (which is frozen during training) for 20 epochs with a batch size of 32 on a single NVIDIA-V100 GPU. We use SGD with an initial learning rate of $8e-3$, a cosine decay schedule and momentum of 0.9 similar to recent work \cite{caron2021emerging, qian2020spatiotemporal}; \textbf{b) Fine-tuning:} We replace the projection head over the SVT with a randomly initialized linear layer, initialize the SVT backbone with our pre-trained weights, and train the network end-to-end for 15 epochs with a batch size of 32 across 4 NVIDIA-V100 GPUs.  We use SGD with a learning rate of $5e-3$ decayed by a factor of 10 at epochs 11 and 14, momentum of 0.9, and weight decay of $1e-4$ following \cite{gberta_2021_ICML}.

During both training of linear classifier and fine-tuning of SVT, we sample two clips of varying spatiotemporal resolution from each video. During evaluation, we follow our proposed slow-fast inference strategy (\cref{subsec:sf_inf}). We use two clips per video sampled at different spatiotemporal resolutions
$(T, W, H) \in \{(8, 224, 224), (64, 96, 96)\}$ 
with 3 spatial crops each for testing (6 clips in total). This is computationally more efficient in comparison to recent works \cite{qian2020spatiotemporal, recasens2021broaden} that uniformly sample 10 clips at similar or high resolutions from full-length videos with 3 crops each for testing (total of 30 clips per video).

\subsection{Results}
We compare SVT with state-of-the-art approaches (trained on RGB input modality for fair comparison) for the downstream task of action recognition.

\vspace{0.1em}
\noindent\textbf{UCF-101 \& HMDB-51:} Our method out-performs state-of-the-art for UCF-101 and is on-par for HMDB-51 (\cref{tbl:sota_comp}). 
While CORP \cite{Hu_2021_ICCV} exhibits higher performance on HMDB-51, we highlight how SVT: \textbf{a)} is pretrained for a much shorter duration (20 epochs) with smaller batch-sizes (32); \textbf{b)} uses a single architectural design across all tasks. CORP \cite{Hu_2021_ICCV} models are pre-trained for 800 epochs with a batch-size of 1024 using 64 NVIDIA-V100 GPUs and uses different variants (CORP$_f$ and CORP$_m$) to obtain optimal performance on different datasets.

\vspace{0.1em}
\noindent\textbf{Kinetics-400:} 
We present our results on Kinetics-400 \cite{kinetics400} in \cref{tbl:sota_k400} where our approach obtains state-of-the-art for both linear evaluation and fine-tuning settings. Performance on Kinetics-400 is heavily dependent on appearance attributes, \ie a large proportion of its videos can
be recognized by a single frame  \cite{zhu_arxiv2020_comprehensiveVideo}. 
Strong performance of SVT on this dataset exhibits how well our proposed approach 
learns appearance related contextual information.

\vspace{0.1em}
\noindent\textbf{SSv2:} Similarly, we obtain state-of-the-art results on SSv2 dataset \cite{goyal2017something} for both linear evaluation and fine-tune settings as presented in \cref{tbl:sota_ssv2}.  Multiple classes in SSv2 share similar backgrounds and object appearance, with complex movements differentiating them \cite{Hu_2021_ICCV}. Performance on this dataset indicates how 
SVT feature representations capture strong motion related contextual cues as well.

\subsection{Ablative Analysis}
We systematically dissect the contribution of each component of our method. 
We study the effect of five individual elements: 
\textbf{a)} different combinations of local and global view correspondences;
\textbf{b)} varying field of view along temporal vs spatial dimensions;
\textbf{c)} temporal sampling strategy;
\textbf{d)} spatial augmentations;
\textbf{e)} slow-fast inference.
In all our ablative experiments, SVT self-supervised training uses a subset of the Kinetics-400 train set containing 60K videos. Evaluation is carried out over \emph{alternate} train-set splits of UCF-101 and HMDB-51. We train SVT for 20 epochs and evaluate using the same setup as described in \cref{subsec:exp_setup}.

\vspace{0.1em}
\noindent\textbf{View Correspondences.}
Learning correspondences between local and global views is the key motivation behind our proposed cross-view correspondences. Since multiple local-global view combinations can be considered for matching and prediction between views, we explore the effect of predicting each type of view from the other in \cref{tbl:ablation_correspondences}. We observe that jointly predicting local to global and global to global view correspondences results in the optimal performance, while predicting global to local or local to local views leads to reduced performance. We believe this trend exists due to the emphasis on learning rich context in the case of joint prediction, which is absent for individual cases. 
{Further, the performance drop for local to local correspondences (non-overlapping views) conforms with previous findings on the effectiveness of temporally closer positive views for contrastive self-supervised losses \cite{qian2020spatiotemporal, Feichtenhofer_large}.}

\vspace{0.1em}
\noindent\textbf{Spatial vs Temporal Field of View.} The optimal combination of spatiotemporal views in \cref{tbl:ablation_correspondences} involves varying the field of view (crops) along both spatial and temporal dimensions (Sec.~\ref{subsec:cross_view_corr}). We study the effects of these variations (spatial or temporal) in  \cref{tbl:ablation_st}. No variation along the spatial dimension denotes that all frames 
are of fixed spatial resolution $224\times 224$ 
with no spatial cropping, 
and no temporal variation denotes that all frames in our views are sampled from a fixed time-axis region of a video. We observe that temporal variations have a significant effect on UCF-101, while variations in the field of view along both spatial and temporal dimension perform the best (\cref{tbl:ablation_st}).

\vspace{0.1em}
\noindent\textbf{Temporal Sampling Strategy.} 
We study how our proposed temporal sampling strategy for motion correspondences (MC) could be replaced with alternate sampling approaches. To verify the effectiveness of MC, we replace it within SVT with an alternate approach. The temporal interval sampling (TIS) strategy in \cite{qian2020spatiotemporal} obtains state-of-the-art performance under their contrastive video self-supervised setting with CNN backbones. Our experiments incorporating TIS in SVT (\cref{tbl:ablation_temporal_aug}) highlight the advantage of our proposed MC sampling strategy over TIS.  

\vspace{0.1em}
\noindent\textbf{Augmentations}: We next explore standard spatial augmentations used on videos.
Temporally consistent augmentations (TCA) proposed in \cite{qian2020spatiotemporal} lead to improvements in their CNN based video self-supervision approach. We evaluate its effect on our approach in \cref{tbl:ablation_aug} which shows slight improvements. Given these performance gains, we adopt TCA in our SVT training process as well.

\vspace{0.1em}
\noindent\textbf{Slow-Fast Inference:} Finally, we study the effect of our proposed Slow-Fast inference (\cref{subsec:sf_inf}) in \cref{tbl:ablation_inf}. We observe higher gains on  HMDB-51 \cite{kuehne2011hmdb}, where the classes are easier to separate with motion information \cite{han2020self}. 

\section{Conclusion}
\label{sec:conclusion}
We present a video transformer based model trained using self-supervised objectives named SVT. Given an input video sequence, our approach first creates a set of spatiotemporally varying views sampled at different scales and frame rates from the same video. We then define two sets of correspondence learning tasks which seek to model the motion properties and cross-view relationships between the sampled clips. Specifically, our self-supervised objective reconstructs one view from the other in the latent space of student and teacher networks. Our approach is fast to train (converges within only a few iterations), does not require negative samples and large batch sizes that are required by previous contrastive video representation  learning methods. Additionally, our SVT  allows modeling long-range spatiotemporal dependencies and can perform dynamic slow-fast inference within a single architecture. SVT is evaluated on four benchmark action recognition datasets where it performs well in comparison to existing state of the art. 

\textbf{Limitations:}
In this work, we explore SVT within the context of RGB input modality. Given large-scale multi-modal video datasets, the additional supervision available in the form of alternate modalities is not used by our current approach. In future work, we will explore how SVT can be modified to utilize multi-modal data sources.


{\small
\bibliographystyle{ieee_fullname}
\bibliography{egbib}

\begin{thebibliography}{10}\itemsep=-1pt

\bibitem{youtube8M}
Sami Abu-El-Haija, Nisarg Kothari, Joonseok Lee, Apostol~(Paul) Natsev, George
  Toderici, Balakrishnan Varadarajan, and Sudheendra Vijayanarasimhan.
\newblock Youtube-8m: A large-scale video classification benchmark.
\newblock In {\em ArXiv preprint}, 2016.

\bibitem{Agrawal_2015_ICCV}
Pulkit Agrawal, Joao Carreira, and Jitendra Malik.
\newblock Learning to see by moving.
\newblock In {\em ICCV}, 2015.

\bibitem{alwassel2019self}
Humam Alwassel, Dhruv Mahajan, Lorenzo Torresani, Bernard Ghanem, and Du Tran.
\newblock Self-supervised learning by cross-modal audio-video clustering.
\newblock In {\em NeurIPS}, 2020.

\bibitem{arnab2021vivit}
Anurag Arnab, Mostafa Dehghani, Georg Heigold, Chen Sun, Mario Lu{\v{c}}i{\'c},
  and Cordelia Schmid.
\newblock Vivit: A video vision transformer.
\newblock {\em ICCV}, 2021.

\bibitem{asano2020selflabelling}
Yuki~Markus Asano, Christian Rupprecht, and Andrea Vedaldi.
\newblock Self-labelling via simultaneous clustering and representation
  learning.
\newblock In {\em ICLR}, 2020.

\bibitem{AMDIM}
Philip Bachman, R~Devon Hjelm, and William Buchwalter.
\newblock Learning representations by maximizing mutual information across
  views.
\newblock In {\em NeurIPS}, 2019.

\bibitem{bautista2016cliquecnn}
Miguel~A. Bautista, Artsiom Sanakoyeu, Ekaterina Sutter, and Björn Ommer.
\newblock Cliquecnn: Deep unsupervised exemplar learning.
\newblock In {\em NeurIPS}, 2016.

\bibitem{Behrmann2021LongSV}
Nadine Behrmann, Mohsen Fayyaz, Juergen Gall, and Mehdi Noroozi.
\newblock Long short view feature decomposition via contrastive video
  representation learning.
\newblock In {\em ICCV}, 2021.

\bibitem{gberta_2021_ICML}
Gedas Bertasius, Heng Wang, and Lorenzo Torresani.
\newblock Is space-time attention all you need for video understanding?
\newblock In {\em ICML}, July 2021.

\bibitem{carion2020end}
Nicolas Carion, Francisco Massa, Gabriel Synnaeve, Nicolas Usunier, Alexander
  Kirillov, and Sergey Zagoruyko.
\newblock End-to-end object detection with transformers.
\newblock In {\em ECCV}, 2020.

\bibitem{caron2018DeepCF}
Mathilde Caron, Piotr Bojanowski, Armand Joulin, and Matthijs Douze.
\newblock Deep clustering for unsupervised learning of visual features.
\newblock In {\em ECCV}, 2018.

\bibitem{caron2020unsupervised}
Mathilde Caron, Ishan Misra, Julien Mairal, Priya Goyal, Piotr Bojanowski, and
  Armand Joulin.
\newblock Unsupervised learning of visual features by contrasting cluster
  assignments.
\newblock In {\em NeurIPS}, 2020.

\bibitem{caron2021emerging}
Mathilde Caron, Hugo Touvron, Ishan Misra, Herv\'e J\'egou, Julien Mairal,
  Piotr Bojanowski, and Armand Joulin.
\newblock Emerging properties in self-supervised vision transformers.
\newblock In {\em ICCV}, 2021.

\bibitem{kinetics400}
Joao Carreira and Andrew Zisserman.
\newblock Quo vadis, action recognition? {A} new model and the {Kinetics}
  dataset.
\newblock In {\em CVPR}, 2017.

\bibitem{chen2021rspnet}
Peihao Chen, Deng Huang, Dongliang He, Xiang Long, Runhao Zeng, Shilei Wen,
  Mingkui Tan, and Chuang Gan.
\newblock Rspnet: Relative speed perception for unsupervised video
  representation learning.
\newblock In {\em AAAI}, volume~1, 2021.

\bibitem{chen2020simple}
Ting Chen, Simon Kornblith, Mohammad Norouzi, and Geoffrey Hinton.
\newblock A simple framework for contrastive learning of visual
  representations.
\newblock In {\em ICML}, 2020.

\bibitem{chen2020improved}
Xinlei Chen, Haoqi Fan, Ross Girshick, and Kaiming He.
\newblock Improved baselines with momentum contrastive learning.
\newblock {\em ArXiv preprint}, 2020.

\bibitem{chen2021mocov3}
Xinlei Chen*, Saining Xie*, and Kaiming He.
\newblock An empirical study of training self-supervised vision transformers.
\newblock {\em ArXiv preprint}, 2021.

\bibitem{chen2021empirical}
Xinlei Chen, Saining Xie, and Kaiming He.
\newblock An empirical study of training self-supervised visual transformers.
\newblock {\em ArXiv preprint}, 2021.

\bibitem{Dave2021TCLRTC}
Ishan~Rajendra Dave, Rohit Gupta, Mamshad~Nayeem Rizve, and Mubarak Shah.
\newblock {TCLR}: Temporal contrastive learning for video representation.
\newblock {\em ArXiv preprint}, 2021.

\bibitem{devlin2018bert}
Jacob Devlin, Ming-Wei Chang, Kenton Lee, and Kristina Toutanova.
\newblock Bert: Pre-training of deep bidirectional transformers for language
  understanding.
\newblock {\em ArXiv preprint}, 2018.

\bibitem{hjelm2020representation}
R Devon et~al.
\newblock Representation learning with video deep infomax.
\newblock {\em ArXiv preprint}, 2020.

\bibitem{Diba_2021_ICCV}
Ali Diba, Vivek Sharma, Reza Safdari, Dariush Lotfi, Saquib Sarfraz, Rainer
  Stiefelhagen, and Luc Van~Gool.
\newblock Vi2clr: Video and image for visual contrastive learning of
  representation.
\newblock In {\em ICCV}, 2021.

\bibitem{doersch2015unsupervised}
Carl Doersch, Abhinav Gupta, and Alexei~A Efros.
\newblock Unsupervised visual representation learning by context prediction.
\newblock In {\em ICCV}, 2015.

\bibitem{doersch2017multi}
Carl Doersch and Andrew Zisserman.
\newblock Multi-task self-supervised visual learning.
\newblock In {\em ICCV}, 2017.

\bibitem{dosovitskiy2020image}
Alexey Dosovitskiy, Lucas Beyer, Alexander Kolesnikov, Dirk Weissenborn,
  Xiaohua Zhai, Thomas Unterthiner, Mostafa Dehghani, Matthias Minderer, Georg
  Heigold, Sylvain Gelly, et~al.
\newblock An image is worth 16x16 words: Transformers for image recognition at
  scale.
\newblock {\em ICLR}, 2021.

\bibitem{dosovitskiy2014discriminative}
Alexey Dosovitskiy, Jost~Tobias Springenberg, Martin Riedmiller, and Thomas
  Brox.
\newblock Discriminative unsupervised feature learning with convolutional
  neural networks.
\newblock In {\em NeurIPS}, 2014.

\bibitem{fan2020pyslowfast}
Haoqi Fan, Yanghao Li, Bo Xiong, Wan-Yen Lo, and Christoph Feichtenhofer.
\newblock Pyslowfast.
\newblock \url{https://github.com/facebookresearch/slowfast}, 2020.

\bibitem{fan2021multiscale}
Haoqi Fan, Bo Xiong, Karttikeya Mangalam, Yanghao Li, Zhicheng Yan, Jitendra
  Malik, and Christoph Feichtenhofer.
\newblock Multiscale vision transformers.
\newblock {\em ICCV}, 2021.

\bibitem{feichtenhofer2019slowfast}
Christoph Feichtenhofer, Haoqi Fan, Jitendra Malik, and Kaiming He.
\newblock Slowfast networks for video recognition.
\newblock In {\em ICCV}, pages 6202--6211, 2019.

\bibitem{Feichtenhofer_large}
Christoph Feichtenhofer, Haoqi Fan, Bo Xiong, Ross Girshick, and Kaiming He.
\newblock A large-scale study on unsupervised spatiotemporal representation
  learning.
\newblock {\em ArXiv preprint}, 2021.

\bibitem{Goroshin_2015_ICCV}
Ross Goroshin, Joan Bruna, Jonathan Tompson, David Eigen, and Yann LeCun.
\newblock Unsupervised learning of spatiotemporally coherent metrics.
\newblock In {\em ICCV}, 2015.

\bibitem{goyal2017something}
Raghav Goyal, Samira~Ebrahimi Kahou, Vincent Michalski, Joanna Materzyńska,
  Susanne Westphal, Heuna Kim, Valentin Haenel, Ingo Fruend, Peter Yianilos,
  Moritz Mueller-Freitag, Florian Hoppe, Christian Thurau, Ingo Bax, and Roland
  Memisevic.
\newblock The "something something" video database for learning and evaluating
  visual common sense.
\newblock In {\em ArXiv preprint}, 2017.

\bibitem{grill2020bootstrap}
Jean-Bastien Grill, Florian Strub, Florent Altch{\'e}, Corentin Tallec,
  Pierre~H Richemond, Elena Buchatskaya, Carl Doersch, Bernardo~Avila Pires,
  Zhaohan~Daniel Guo, Mohammad~Gheshlaghi Azar, et~al.
\newblock Bootstrap your own latent: A new approach to self-supervised
  learning.
\newblock In {\em NeurIPS}, 2020.

\bibitem{han2019video}
Tengda Han, Weidi Xie, and Andrew Zisserman.
\newblock Video representation learning by dense predictive coding.
\newblock In {\em ICCV}, 2019.

\bibitem{han2020self}
Tengda Han, Weidi Xie, and Andrew Zisserman.
\newblock Self-supervised co-training for video representation learning.
\newblock {\em NeurIPS}, 2020.

\bibitem{he2020momentum}
Kaiming He, Haoqi Fan, Yuxin Wu, Saining Xie, and Ross Girshick.
\newblock Momentum contrast for unsupervised visual representation learning.
\newblock In {\em CVPR}, 2020.

\bibitem{henaff2019data}
Olivier~J H{\'e}naff, Ali Razavi, Carl Doersch, SM Eslami, and Aaron van~den
  Oord.
\newblock Data-efficient image recognition with contrastive predictive coding.
\newblock In {\em ICML}, 2020.

\bibitem{Hu_2021_ICCV}
Kai Hu, Jie Shao, Yuan Liu, Bhiksha Raj, Marios Savvides, and Zhiqiang Shen.
\newblock Contrast and order representations for video self-supervised
  learning.
\newblock In {\em ICCV}, 2021.

\bibitem{Huang_2021_ICCV}
Deng Huang, Wenhao Wu, Weiwen Hu, Xu Liu, Dongliang He, Zhihua Wu, Xiangmiao
  Wu, Mingkui Tan, and Errui Ding.
\newblock Ascnet: Self-supervised video representation learning with
  appearance-speed consistency.
\newblock In {\em ICCV}, 2021.

\bibitem{huang2019unsupervised}
Jiabo Huang, Qi Dong, Shaogang Gong, and Xiatian Zhu.
\newblock Unsupervised deep learning by neighbourhood discovery.
\newblock In {\em ICML}, 2019.

\bibitem{Huang2021SSVR}
Lianghua Huang, Yu Liu, Bin Wang, Pan Pan, Yinghui Xu, and Rong Jin.
\newblock Self-supervised video representation learning by context and motion
  decoupling.
\newblock {\em CVPR}, 2021.

\bibitem{DBLP:journals/corr/IsolaZKA15}
Phillip Isola, Daniel Zoran, Dilip Krishnan, and Edward~H. Adelson.
\newblock Learning visual groups from co-occurrences in space and time.
\newblock 2016.

\bibitem{Jenni_2021_ICCV}
Simon Jenni and Hailin Jin.
\newblock Time-equivariant contrastive video representation learning.
\newblock In {\em ICCV}, 2021.

\bibitem{kahatapitiya2021coarse}
Kumara Kahatapitiya and Michael~S Ryoo.
\newblock Coarse-fine networks for temporal activity detection in videos.
\newblock In {\em CVPR}, 2021.

\bibitem{khan2021transformers}
Salman Khan, Muzammal Naseer, Munawar Hayat, Syed~Waqas Zamir, Fahad~Shahbaz
  Khan, and Mubarak Shah.
\newblock Transformers in vision: A survey.
\newblock {\em ArXiv preprint}, 2021.

\bibitem{kingma15adam}
Diederik~P. Kingma and Jimmy Ba.
\newblock Adam: A method for stochastic optimization.
\newblock In {\em ICLR}, 2015.

\bibitem{komodakis2018unsupervised}
Nikos Komodakis and Spyros Gidaris.
\newblock Unsupervised representation learning by predicting image rotations.
\newblock In {\em ICLR}, 2018.

\bibitem{kuehne2011hmdb}
Hildegard Kuehne, Hueihan Jhuang, Est{\'\i}baliz Garrote, Tomaso Poggio, and
  Thomas Serre.
\newblock {HMDB}: {A} large video database for human motion recognition.
\newblock In {\em ICCV}, 2011.

\bibitem{le2020contrastive}
Phuc~H Le-Khac, Graham Healy, and Alan~F Smeaton.
\newblock Contrastive representation learning: A framework and review.
\newblock {\em IEEE Access}, 2020.

\bibitem{Lin_2021_ICCV}
Yuanze Lin, Xun Guo, and Yan Lu.
\newblock Self-supervised video representation learning with meta-contrastive
  network.
\newblock In {\em ICCV}, 2021.

\bibitem{liu2021swin}
Ze Liu, Yutong Lin, Yue Cao, Han Hu, Yixuan Wei, Zheng Zhang, Stephen Lin, and
  Baining Guo.
\newblock Swin transformer: Hierarchical vision transformer using shifted
  windows.
\newblock {\em ArXiv preprint}, 2021.

\bibitem{liu2021video}
Ze Liu, Jia Ning, Yue Cao, Yixuan Wei, Zheng Zhang, Stephen Lin, and Han Hu.
\newblock Video swin transformer.
\newblock {\em ArXiv preprint}, 2021.

\bibitem{mathieu2015deep}
Michael Mathieu, Camille Couprie, and Yann LeCun.
\newblock Deep multi-scale video prediction beyond mean square error.
\newblock In {\em ICLR}, 2016.

\bibitem{PIRL}
Ishan Misra and Laurens van~der Maaten.
\newblock Self-supervised learning of pretext-invariant representations.
\newblock In {\em CVPR}, 2020.

\bibitem{Misra-2016-5596}
Ishan Misra, C.~Lawrence Zitnick, and Martial Hebert.
\newblock Shuffle and learn: Unsupervised learning using temporal order
  verification.
\newblock In {\em ECCV}, 2016.

\bibitem{naseer2021intriguing}
Muzammal Naseer, Kanchana Ranasinghe, Salman Khan, Munawar Hayat, Fahad~Shahbaz
  Khan, and Ming-Hsuan Yang.
\newblock Intriguing properties of vision transformers.
\newblock {\em ArXiv preprint}, 2021.

\bibitem{noroozi2016unsupervised}
Mehdi Noroozi and Paolo Favaro.
\newblock Unsupervised learning of visual representations by solving jigsaw
  puzzles.
\newblock In {\em ECCV}, 2016.

\bibitem{oord2018representation}
Aaron van~den Oord, Yazhe Li, and Oriol Vinyals.
\newblock Representation learning with contrastive predictive coding.
\newblock {\em NeurIPS}, 2018.

\bibitem{pan2021videomoco}
Tian Pan, Yibing Song, Tianyu Yang, Wenhao Jiang, and Wei Liu.
\newblock Videomoco: Contrastive video representation learning with temporally
  adversarial examples.
\newblock In {\em CVPR}, 2021.

\bibitem{pathak2016context}
Deepak Pathak, Philipp Krahenbuhl, Jeff Donahue, Trevor Darrell, and Alexei~A
  Efros.
\newblock Context encoders: Feature learning by inpainting.
\newblock In {\em CVPR}, 2016.

\bibitem{PatrauceanHC16}
Viorica P{\u a}tr{\u a}ucean, Ankur Handa, and Roberto Cipolla.
\newblock Spatio-temporal video autoencoder with differentiable memory.
\newblock In {\em ICLR (Workshop)}, 2016.

\bibitem{piergiovanni2020evolving}
AJ Piergiovanni, Anelia Angelova, and Michael~S. Ryoo.
\newblock Evolving losses for unsupervised video representation learning.
\newblock In {\em CVPR}, 2020.

\bibitem{qian2020spatiotemporal}
Rui Qian, Tianjian Meng, Boqing Gong, Ming-Hsuan Yang, Huisheng Wang, Serge
  Belongie, and Yin Cui.
\newblock Spatiotemporal contrastive video representation learning.
\newblock {\em CVPR}, 2021.

\bibitem{recasens2021broaden}
Adri{\`a} Recasens, Pauline Luc, Jean-Baptiste Alayrac, Luyu Wang, Florian
  Strub, Corentin Tallec, Mateusz Malinowski, Viorica Patraucean, Florent
  Altch{\'e}, Michal Valko, et~al.
\newblock Broaden your views for self-supervised video learning.
\newblock {\em ICCV}, 2021.

\bibitem{imagenet}
Olga Russakovsky, Jia Deng, Hao Su, Jonathan Krause, Sanjeev Satheesh, Sean Ma,
  Zhiheng Huang, Andrej Karpathy, Aditya Khosla, Michael Bernstein,
  Alexander~C. Berg, and Li Fei-Fei.
\newblock Imagenet large scale visual recognition challenge.
\newblock {\em IJCV}, 2015.

\bibitem{Ryoo2021TokenLearnerWC}
Michael~S. Ryoo, A.~J. Piergiovanni, Anurag Arnab, Mostafa Dehghani, and Anelia
  Angelova.
\newblock Tokenlearner: What can 8 learned tokens do for images and videos?
\newblock {\em ArXiv preprint}, 2021.

\bibitem{sharir2021image}
Gilad Sharir, Asaf Noy, and Lihi Zelnik-Manor.
\newblock An image is worth 16x16 words, what is a video worth?
\newblock {\em ArXiv preprint}, 2021.

\bibitem{soomro2012ucf}
Khurram Soomro, Amir~Roshan Zamir, and Mubarak Shah.
\newblock {UCF101}: {A} dataset of 101 human actions classes from videos in the
  wild.
\newblock {\em ArXiv preprint}, 2012.

\bibitem{pmlr-v37-srivastava15}
Nitish Srivastava, Elman Mansimov, and Ruslan Salakhudinov.
\newblock Unsupervised learning of video representations using lstms.
\newblock In {\em ICML}, 2015.

\bibitem{Steiner2021HowTT}
Andreas Steiner, Alexander Kolesnikov, Xiaohua Zhai, Ross Wightman, Jakob
  Uszkoreit, and Lucas Beyer.
\newblock How to train your vit? data, augmentation, and regularization in
  vision transformers.
\newblock {\em ArXiv preprint}, 2021.

\bibitem{tarvainen2017mean}
Antti Tarvainen and Harri Valpola.
\newblock Mean teachers are better role models: Weight-averaged consistency
  targets improve semi-supervised deep learning results.
\newblock {\em ArXiv preprint}, 2017.

\bibitem{tian2017deepcluster}
Kai Tian, Shuigeng Zhou, and Jihong Guan.
\newblock Deepcluster: A general clustering framework based on deep learning.
\newblock In {\em ECML/PKDD}, 2017.

\bibitem{tian2019contrastive}
Yonglong Tian, Dilip Krishnan, and Phillip Isola.
\newblock Contrastive multiview coding.
\newblock {\em ECCV}, 2020.

\bibitem{tian2020makes}
Yonglong Tian, Chen Sun, Ben Poole, Dilip Krishnan, Cordelia Schmid, and
  Phillip Isola.
\newblock What makes for good views for contrastive learning.
\newblock In {\em NeurIPS}, 2020.

\bibitem{pmlr-v139-touvron21a}
Hugo Touvron, Matthieu Cord, Matthijs Douze, Francisco Massa, Alexandre
  Sablayrolles, and Herve Jegou.
\newblock Training data-efficient image transformers and distillation through
  attention.
\newblock In {\em ICML}, 2021.

\bibitem{vaswani2017attention}
Ashish Vaswani, Noam Shazeer, Niki Parmar, Jakob Uszkoreit, Llion Jones,
  Aidan~N Gomez, Lukasz Kaiser, and Illia Polosukhin.
\newblock Attention is all you need.
\newblock {\em NeurIPS}, 2017.

\bibitem{vincent2008extracting}
Pascal Vincent, Hugo Larochelle, Yoshua Bengio, and Pierre-Antoine Manzagol.
\newblock Extracting and composing robust features with denoising autoencoders.
\newblock In {\em ICML}, 2008.

\bibitem{Vondrick16a}
Carl Vondrick, Hamed Pirsiavash, and Antonio Torralba.
\newblock Generating videos with scene dynamics.
\newblock In {\em NeurIPS}, 2016.

\bibitem{vondrick2018tracking}
Carl Vondrick, Abhinav Shrivastava, Alireza Fathi, Sergio Guadarrama, and Kevin
  Murphy.
\newblock Tracking emerges by colorizing videos.
\newblock In {\em ECCV}, 2018.

\bibitem{walker2016uncertain}
Jacob Walker, Carl Doersch, Abhinav Gupta, and Martial Hebert.
\newblock An uncertain future: Forecasting from static images using variational
  autoencoders.
\newblock In {\em ECCV}, 2016.

\bibitem{wang2021removing}
Jinpeng Wang, Yuting Gao, Ke Li, Yiqi Lin, Andy~J Ma, Hao Cheng, Pai Peng,
  Rongrong Ji, and Xing Sun.
\newblock Removing the background by adding the background: Towards background
  robust self-supervised video representation learning.
\newblock In {\em CVPR}, 2021.

\bibitem{wang2018non}
Xiaolong Wang, Ross Girshick, Abhinav Gupta, and Kaiming He.
\newblock Non-local neural networks.
\newblock In {\em Proceedings of the IEEE conference on computer vision and
  pattern recognition}, pages 7794--7803, 2018.

\bibitem{7410677}
Xiaolong Wang and Abhinav Gupta.
\newblock Unsupervised learning of visual representations using videos.
\newblock In {\em ICCV}.

\bibitem{wang2020end}
Yuqing Wang, Zhaoliang Xu, Xinlong Wang, Chunhua Shen, Baoshan Cheng, Hao Shen,
  and Huaxia Xia.
\newblock End-to-end video instance segmentation with transformers.
\newblock {\em ArXiv preprint}, 2020.

\bibitem{rw2019timm}
Ross Wightman.
\newblock Pytorch image models.
\newblock \url{https://github.com/rwightman/pytorch-image-models}, 2019.

\bibitem{xiao2021modist}
Fanyi Xiao, Joseph Tighe, and Davide Modolo.
\newblock Modist: Motion distillation for self-supervised video representation
  learning.
\newblock {\em ArXiv preprint}, 2021.

\bibitem{xie2016unsupervised}
Junyuan Xie, Ross Girshick, and Ali Farhadi.
\newblock Unsupervised deep embedding for clustering analysis.
\newblock In {\em ICML}, 2016.

\bibitem{zhang2020dynamic}
Li Zhang, Dan Xu, Anurag Arnab, and Philip~HS Torr.
\newblock Dynamic graph message passing networks.
\newblock In {\em CVPR}, 2020.

\bibitem{zhang2016colorful}
Richard Zhang, Phillip Isola, and Alexei~A Efros.
\newblock Colorful image colorization.
\newblock In {\em ECCV}, 2016.

\bibitem{zhu_arxiv2020_comprehensiveVideo}
Yi Zhu, Xinyu Li, Chunhui Liu, Mohammadreza Zolfaghari, Yuanjun Xiong, Chongruo
  Wu, Zhi Zhang, Joseph Tighe, R Manmatha, and Mu Li.
\newblock A comprehensive study of deep video action recognition.
\newblock {\em ArXiv preprint}, 2020.

\end{thebibliography}
}

\newpage
\appendix

\setcounter{table}{0}
\setcounter{figure}{0}

\renewcommand{\thetable}{\Roman{table}}
\renewcommand{\thefigure}{\Roman{figure}}

\begin{figure*}[t]
	\centering
	\includegraphics[width=0.12\linewidth]{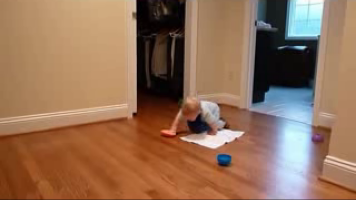}
	\includegraphics[width=0.12\linewidth]{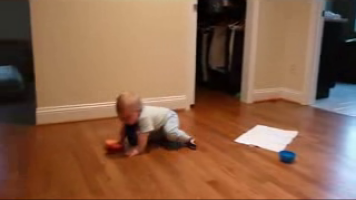}
	\includegraphics[width=0.12\linewidth]{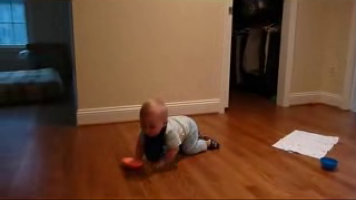}
	\includegraphics[width=0.12\linewidth]{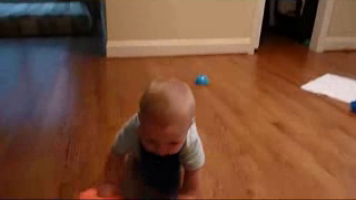}
	\hspace{0.1em}
	\includegraphics[width=0.12\linewidth]{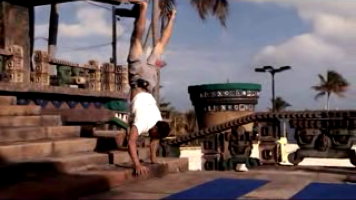}
	\includegraphics[width=0.12\linewidth]{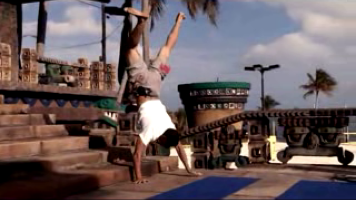}
	\includegraphics[width=0.12\linewidth]{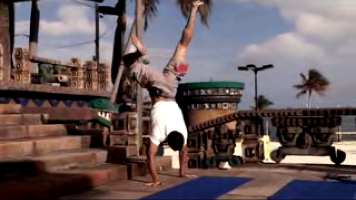}
	\includegraphics[width=0.12\linewidth]{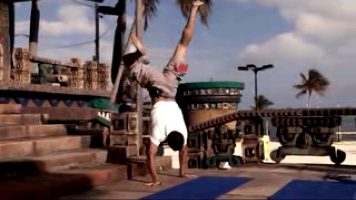}

	\includegraphics[width=0.12\linewidth]{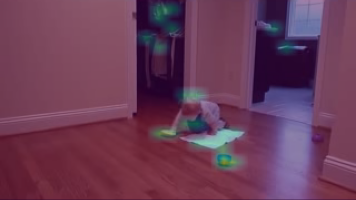}
	\includegraphics[width=0.12\linewidth]{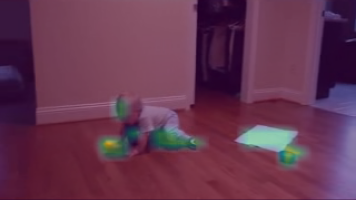}
	\includegraphics[width=0.12\linewidth]{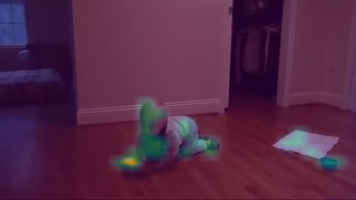}
	\includegraphics[width=0.12\linewidth]{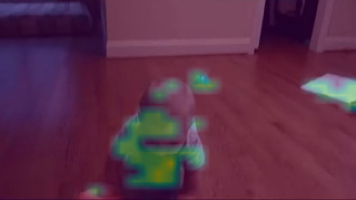} \hspace{0.1em}
	\includegraphics[width=0.12\linewidth]{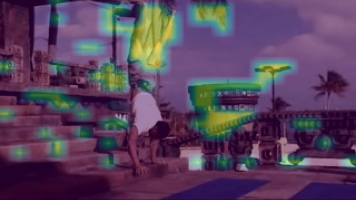}
	\includegraphics[width=0.12\linewidth]{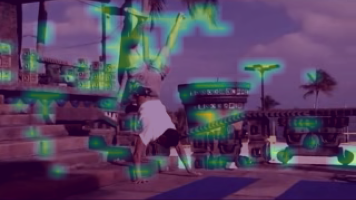}
	\includegraphics[width=0.12\linewidth]{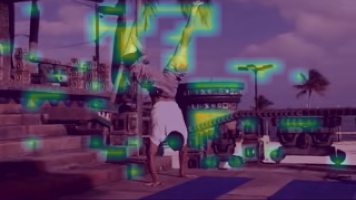}
	\includegraphics[width=0.12\linewidth]{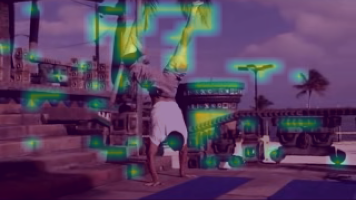}
	
	\includegraphics[width=0.12\linewidth]{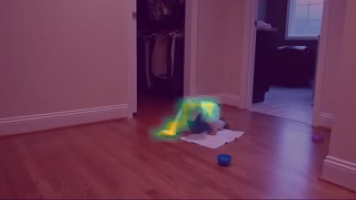}
	\includegraphics[width=0.12\linewidth]{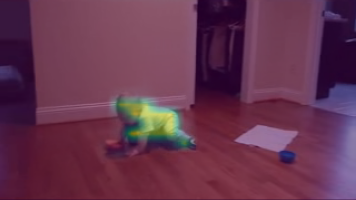}
	\includegraphics[width=0.12\linewidth]{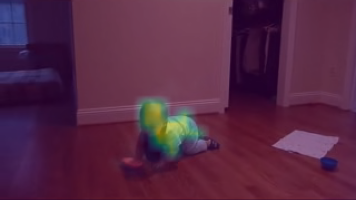}
	\includegraphics[width=0.12\linewidth]{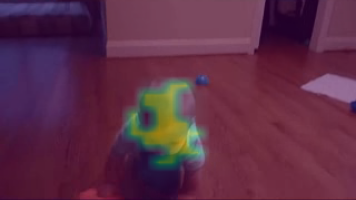}
    \hspace{0.1em}
    \includegraphics[width=0.12\linewidth]{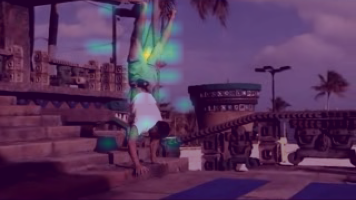}
	\includegraphics[width=0.12\linewidth]{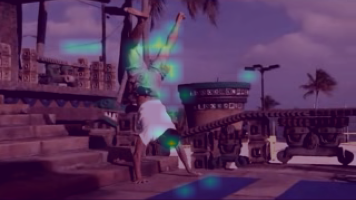}
	\includegraphics[width=0.12\linewidth]{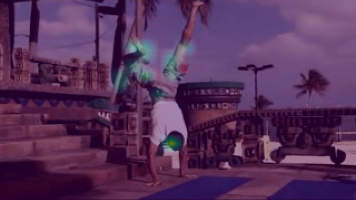}
	\includegraphics[width=0.12\linewidth]{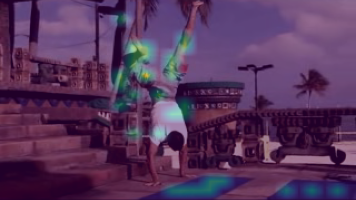}
	
	\caption{\textbf{Attention Visualization:} We uniformly sample four frames from two videos (cols 1-4 and 5-8 respectively) and visualize the attention from the classification token of self-supervised vision transformer DINO \cite{caron2021emerging} (second row) 
	and our SVT (last row). Observe how DINO attention is scattered around multiple objects, while 
	SVT is focused on \textit{`crawling baby'} and \textit{`person walking on hands'} across frames  which are the salient objects for these action. This highlights how SVT learns to pay attention to the motion within a video.
	}
	\label{fig:attn_vis}
\end{figure*}



\section{View Routing and Matching}
\label{app:view_routes}

In the main paper (Sec.~\ref{subsec:training}), we illustrate the concept of SVT using a single global view passed through the teacher, which generates target for all the other views passed through the student model. However, in practice,  multiple global views are all passed through the teacher model, and we separately map each student view (global and local) to the multiple teacher targets. In the case of two global views, $g1$ ($T=8$) and $g2$ ($T=16$), we obtain two targets, $\Tilde{f}_{gt}^{(1)}$ and $\Tilde{f}_{gt}^{(2)}$. Both these global views are also passed through the student model to obtain $\Tilde{f}_{gs}^{(1)}$ and $\Tilde{f}_{gs}^{(2)}$. We map $\Tilde{f}_{gs}^{(1)}$ to $\Tilde{f}_{gt}^{(2)}$ and $\Tilde{f}_{gs}^{(2)}$ to $\Tilde{f}_{gt}^{(1)}$. The local views passed through the student that generates $\Tilde{f}_{ls}^{(1)} ... \Tilde{f}_{gs}^{(8)}$ which are separately mapped to both teacher targets, $\Tilde{f}_{gt}^{(1)}$ and $\Tilde{f}_{gt}^{(2)}$. Our proposed loss is applied over each mapped student-teacher feature pair.

\section{Comparison to Supervised Training}
In SVT, we use a standard ViT backbone with split attention across space and time dimensions similar to \cite{gberta_2021_ICML}. We compare SVT with supervised pre-training based initialization for Kinetics-400 training reported in \cite{gberta_2021_ICML}. For fairness, our comparison with \cite{gberta_2021_ICML} includes the highest reported input resolution used in their work since the SVT uses slow-fast inference. These results are presented in \Cref{tbl:sup_comp}.

\begin{table}[h]
\caption{Comparison of SVT with supervised pretraining methods containing similar backbone (ViT-B) on Kinetics-400. For each different pre-training strategy, we finetune on Kinetics-400 and report accuracy (top-1) on Kinetics-400 validation set.}
\vspace{-0.5em}
\centering
\scalebox{1}{
\begin{tabular}{l|c|c}
\toprule
\rowcolor{Gray} 
Pretrain Dataset      & Supervision & Accuracy \\ \midrule
Random-init  & -          & 64.8 \cite{gberta_2021_ICML} \\ 
ImageNet-1K  & \cmark     & 75.8 \cite{gberta_2021_ICML} \\ 
ImageNet-21K & \cmark     & 79.7 \cite{gberta_2021_ICML} \\ \midrule
ImageNet-1K  & \xmark     & 69.9     \\ 
Kinetics-400  & \xmark    & 78.1    \\ 
\bottomrule
\end{tabular}
}
\label{tbl:sup_comp}
\vspace{-0.5em}
\end{table}

\section{Dataset Description}

We use the Kinetics-400 \cite{kinetics400} training set for the SVT self-supervised training and its validation set for evaluation of learned self-supervised representations. 
Kinetics-400 is a large-scale dataset containing 240k training videos and 20k validation videos belonging to 400 different action classes. 
On average, these videos are of duration around 10 seconds, with 25 frames per second (\ie, around 250 frames per video). Interestingly, most classes of this dataset are considered to be separable with appearance information alone \cite{zhu_arxiv2020_comprehensiveVideo}.
In addition to Kinetics-400, we evaluate our approach on three downstream datasets, UCF-101 \cite{soomro2012ucf}, HMBD-51 \cite{kuehne2011hmdb}, and Something-Something v2 (SSv2) \cite{goyal2017something}. UCF-101 and  HMBD-51 are small-scale datasets each containing 13k videos (9.5k/3.7k train/test) belonging to 101 classes and 5k (3.5k/1.5k train/test) videos belonging to 51 classes respectively, while SSv2 is a large-scale dataset heavily focused on motion with 168k training and 24k validation videos belonging to 174 action classes. Unlike UCF101 and HMDB51 which contain action classes similar to Kinetics-400, SSv2 contains very different actions involving complex human object interactions, such as `Moving something up' or `Pushing something from left to right'.

\section{Future Directions}
As discussed in the main paper, the key limitation of SVT is being constrained to operating within a single modality input (RGB video). We hope to explore how SVT to can improved to utilize alternate modalities (Optical Flow, Audio) for better self supervision in future work. 

In this work, we focus on evaluating the effectiveness of our proposed cross-view and motion correspondences (that compose the core of SVT) in relation to ViT backbones. The question of applicability of our proposed approach under convolutional neural network (CNN) settings remains unexplored. However, we highlight that the  main components (temporal attention, dynamic input sizes, and slow-fast inference) of our proposed SVT are designed to leverage some unique characteristics of ViTs, which could not be directly implemented with a CNN backbone. On the other hand, we believe that self-distillation and view matching, also core to SVT, can be applied to CNNs and is an interesting future direction.

Another key issue is the significant drop in performance (top-1 accuracy) for linear evaluation in large-scale datasets (Kinetics-400 and SSv2). Particularly in SSv2, our features perform poorly in the linear evaluation setting (in comparison to fine-tune setting). A key reason for this could be the significant domain different between Kinetics-400 and SSv2 (as opposed to UCF-101 and HMDB-51 which contain videos and classes more similar to Kinetics-400). The self-supervised training phase of SVT uses Kinetics-400 only, and the SSv2 experiments use that representation for linear evaluation. An interesting future direction we hope to explore is self-supervised training using the SSv2 dataset itself, which could potentially reveal more interesting insights on representations learned by SVT.

\section{Attention Visualization}

Following the approach in \cite{caron2021emerging}, we visualize the attention of our classification token (feature vector) towards each spatiotemporal patch token within the final encoder block of SVT for two randomly selected videos. As illustrated in \cref{fig:attn_vis}, SVT attends to the regions of motion in these videos, even in the case of highly detailed backgrounds (right). Attention to the salient moving object in each case qualitatively demonstrates how our proposed cross-view and motion correspondences learn spatiotemporally invariant representations.

\end{document}